\documentclass[letterpaper, 10 pt, conference]{ieeeconf}
\IEEEoverridecommandlockouts
\usepackage[utf8]{inputenc}
\usepackage[noend]{algpseudocode}
\usepackage{amsmath}
\usepackage{amssymb}    
\usepackage[utf8]{inputenc}
\usepackage{xcolor}
\usepackage{graphicx}
\usepackage{hyperref}
\usepackage{booktabs}
\usepackage{color}
\usepackage{array}
\usepackage{booktabs}
\usepackage{multirow}
\usepackage{hyperref}
\usepackage{makecell}

\usepackage{siunitx}
\usepackage[space]{cite}
\usepackage[linesnumbered,boxed,ruled,commentsnumbered]{algorithm2e}
\graphicspath{{figures/}}
\DeclareGraphicsExtensions{.pdf,.png,.jpg,.eps}
\usepackage{tabularx}
\usepackage{threeparttable} 
\newcolumntype{Y}{>{\centering\arraybackslash}X}
\begin{document}
	%
	\title{\LARGE \bf FAST-Calib: LiDAR-Camera Extrinsic Calibration in One Second}

	\author{Chunran Zheng$^{1}$ and Fu Zhang$^{1}$ 
		\thanks{$^{1}$C. Zheng and F. Zhang are with the Department of Mechanical Engineering, The University of Hong Kong, Hong Kong Special Administrative Region, People's Republic of China.
			{\tt\footnotesize $\{$zhengcr$\}$@connect.hku.hk}, {\tt\footnotesize $ $fuzhang$ $@hku.hk}
		}
	}
	
	\markboth{Journal of \LaTeX\ Class Files,~Vol.~6, No.~1, January~2007}%
	{Shell \MakeLowercase{\textit{et al.}}: Bare Demo of IEEEtran.cls for Journals}
	
	\maketitle
	\thispagestyle{empty}
	
	\begin{abstract}
This paper proposes FAST-Calib, a fast and user-friendly LiDAR–camera extrinsic calibration tool based on a custom-made 3D target. FAST-Calib supports both mechanical and solid-state LiDARs by leveraging an efficient and reliable edge extraction algorithm that is agnostic to LiDAR scan patterns. It also compensates for edge dilation artifacts caused by LiDAR spot spread through ellipse fitting, and supports joint optimization across multiple scenes.
We validate FAST-Calib on three LiDAR models (Ouster, Avia, and Mid360), each paired with a wide-angle camera. Experimental results demonstrate superior accuracy and robustness compared to existing methods. With point-to-point registration errors consistently below \SI{6.5}{\milli\meter} and total processing time under \SI{0.7}{\second}, FAST-Calib provides an efficient, accurate, and target-based automatic calibration pipeline.
We have open-sourced our code and dataset on GitHub\footnote[2]{\url{https://github.com/hku-mars/FAST-Calib}} to benefit the robotics community.
	\end{abstract}
	
	\IEEEpeerreviewmaketitle
	
	\section{Introduction}
	Extrinsic calibration between LiDAR and camera aims to estimate the rigid transformation between their spatial coordinate systems. This is a fundamental requirement for multi-sensor fusion in autonomous driving applications such as tracking\cite{dimitrievski2019behavioral}, mapping\cite{zheng2022fast, zheng2024fast, liu2024neural, liu2025gs, hong2025gs}, and object detection \cite{zhao2020fusion, li2022deepfusion}.
    In production lines, calibration is typically performed in dedicated calibration rooms, where controlled environments and standardized procedures ensure high accuracy and repeatability. Compared to scene-based methods \cite{yuan2021pixel, ye2024mfcalib, koide2023general, pandey2012automatic}, such setups yield lower inter-batch variance and more stable results.
    However, when attempting to transfer factory calibration boards to laboratory environments, existing target-based methods (e.g., Velo2Cam\cite{beltran2022automatic}, JointCalib\cite{yan2023joint}) encounter several practical limitations:
    1) \textbf{Limited Compatibility:} Most existing tools cannot support both mechanical and solid-state LiDARs due to their different scan patterns;
    2) \textbf{Low Automation:} 
    In factory settings, calibration boards are placed at fixed positions for easy detection. In lab environments, arbitrary placement requires manual depth filtering to extract the region of interest from the LiDAR point cloud;
    3) \textbf{No Multi-Scene Support:} 
    In factory settings, multiple boards are placed within the sensor Field of View (FoV) to increase robustness. In lab environments, space constraints typically allow only a single board, requiring multiple captures and joint optimization; 
    4) \textbf{Low Efficiency:} Multi-scene calibration often involves multiple processing steps and cannot achieve second-level calibration speed;
    5) \textbf{Edge Dilation Problem:} For LiDARs with large spot sizes (e.g., Livox LiDAR), the edge of calibration boards often appears expanded, leading to inaccurate feature correspondence between LiDAR and camera data.
    To the best of our knowledge, no open-source tool simultaneously addresses all of these challenges. The absence of a practical and general-purpose calibration solution remains a key bottleneck for the deployment of LiDAR-camera fusion.
    To this end, we propose FAST-Calib: a fast and user-friendly extrinsic calibration tool for LiDAR-camera systems (e.g., FAST-LIVO2). Our contributions are as follows:
\begin{enumerate}
  \item Robust extraction of circular hole edges that is agnostic to LiDAR scan patterns, supporting both mechanical and solid-state LiDARs.
  \item Ellipse fitting to compensate for edge dilation caused by LiDAR spot spread, improving the accuracy of hole center extraction.
  \item Providing an efficient, accurate, and target-based automatic calibration pipeline with multi-scene joint refinement, allowing production-line calibration procedures to be replicated in laboratory settings.
\end{enumerate}	

\begin{figure}[t]
	\begin{center}
		{\includegraphics[width=1\columnwidth]{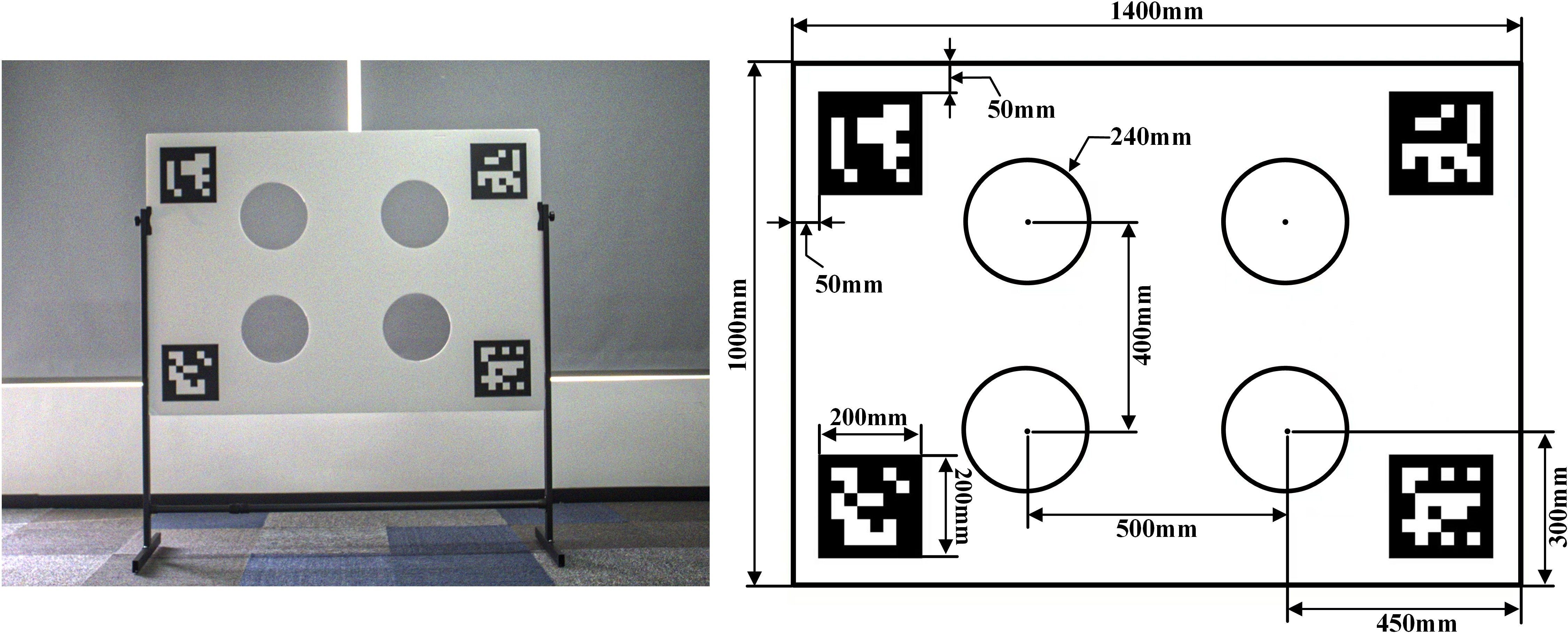}}
	\end{center}
      \vspace{-0.2cm}
	\caption{\label{fig:target}Left: The custom-made calibration target; Right: The corresponding technical drawing with annotated dimensions.}
    \vspace{-0.2cm}
\end{figure}
	\section{Related works}
	LiDAR-camera extrinsic calibration methods are generally divided into target-based and targetless approaches, with the latter including motion-based and scene-based methods.
	
	\subsection{Targetless Calibration\label{targetless}}


Scene-based methods avoid explicit detection of known calibration targets and instead rely on natural features such as planes and edges. For example, \cite{zhu2021camvox} projects LiDAR points onto the image plane, encodes them with depth and reflectivity, and extracts 2D edges from the resulting projection map for matching with image edges. In \cite{pandey2012automatic, koide2023general}, the extrinsic parameters are optimized by maximizing mutual information (NID or MI metrics) between LiDAR reflectivity and image intensity. \cite{yuan2021pixel} also directly extract 3D edge features from the LiDAR point cloud for subsequent matching. 
However, scene-based methods are highly dependent on environmental conditions. They assume that edges caused by photometric changes in the image correspond physically to geometric edges in the LiDAR point cloud. In practice, such conditions are hard to guarantee, which significantly limits the stability and robustness of these methods.

Motion-based methods \cite{tsai1989new, nagy2019online} estimate extrinsic parameters by formulating the problem as hand–eye calibration on a rigid sensor setup. They require hardware time synchronization between the LiDAR and camera, as well as sufficiently excited trajectories. Accurate per-sensor motion estimation is also critical, but often difficult to achieve in practice.
\subsection{Target-based Calibration\label{target-based}}

Compared to targetless methods, the target-based approach offers a more straightforward and environment-independent solution for LiDAR-camera extrinsic calibration, analogous to the well-known checkerboard-based method\cite{zhang2002flexible} for camera intrinsics.
Once the 3D coordinates of points on the calibration target and their corresponding 2D projections in the image are obtained, the transformation between the LiDAR and camera can be estimated directly. 
Some existing methods\cite{unnikrishnan2005fast,zhou2018automatic,cui2020acsc} directly use checkerboards to establish correspondences between the camera and LiDAR. However, these approaches rely heavily on the LiDAR’s reflectivity measurement accuracy. In addition, the strong reflectance contrast between black and white regions often results in noisy or incomplete checkerboard point clouds.
To address these issues, several studies have proposed more practical 3D structured calibration targets, such as Velo2Cam\cite{beltran2022automatic} and JointCalib\cite{yan2023joint}. However, these methods are only applicable to mechanical LiDARs and require labor-intensive manual annotation of the LiDAR point cloud. 
Moreover, due to limited feature extraction accuracy, they often require multiple data captures, resulting in substantial effort. 
In this work, we adopt the calibration target from \cite{beltran2022automatic} but address these limitations by enabling an efficient and automatic calibration pipeline that supports diverse LiDAR types.
\begin{figure*}[t]
	\begin{center}
		{\includegraphics[width=1.85\columnwidth]{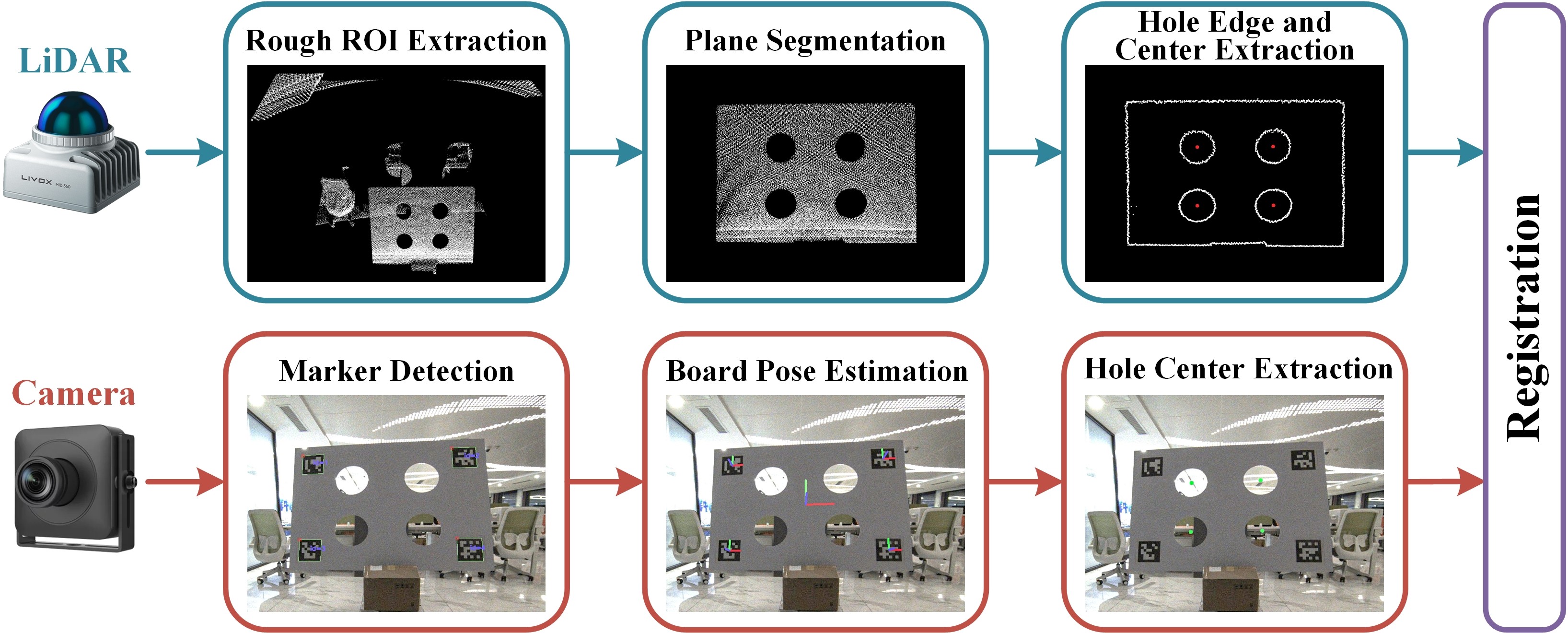}}
	\end{center}
	\vspace{-0.3cm}
	\caption{\label{fig:framework}System overview of FAST-Calib.}
	\vspace{-0.4cm}
\end{figure*}
\section{Methodology}
  
  
	
    \subsection{Overview}\label{subsec:overview}
    To prepare the calibration data, we first place the custom-made board\cite{beltran2022automatic} within the overlapping FoV of both sensors.
    We then merge multiple LiDAR scans to build a dense point cloud. For 16-line mechanical LiDARs, the accumulated data may still be too sparse. In such cases, we adopt the LiDAR odometry mode from FAST-LIVO2, which allows slight pitch movements to aggregate more points. This results in a paired dense point cloud and image for subsequent processing. 
    
    As shown in Fig. \ref{fig:framework}, FAST-Calib consists of two main stages.
    In the LiDAR data processing stage (blue part), we perform rough ROI extraction, plane segmentation and circular hole edge extraction to localize the four hole centers in the LiDAR coordinate frame.
    In the camera data processing stage (red part), we use four ArUco markers\cite{garrido2014automatic} placed at known positions on the calibration target to infer the corresponding hole centers in the camera coordinate frame.
    Finally, the extrinsic transformation is estimated via geometric registration using 4 (or 4$n$) pairs of 3D–3D correspondences.
	\subsection{Camera Data Processing\label{camera_data}}
We first perform ArUco marker detection, which provides geometric references for recovering the positions of the circular holes.
Given the known camera intrinsics and the physical dimensions of the markers, the 3D pose of each marker relative to the camera frame is estimated by solving a Perspective-n-Point (PnP) problem.
The average pose of the four detected markers is then adopted as the estimated pose of the calibration board, thereby determining the position and orientation of the board coordinate frame.
Using the known physical layout of the four hole centers with respect to the board frame, their corresponding 3D positions in the camera frame can be computed.
The resulting hole centers are denoted as the point set $\mathcal{P}_C$.

\subsection{LiDAR Data Processing\label{lidar_data}}

\subsubsection{ROI Extraction\label{roi_extract}}
This section is divided into two stages: rough ROI extraction and plane segmentation.
We begin with rough ROI extraction by applying pass-through filters along the X, Y, and Z axes to isolate the point cloud region containing the calibration target. 
The goal is to roughly filter out most of the points outside the target region, without needing fine-grained tuning of the pass-through parameters.
Note that some points behind or beside the calibration target may remain after filtering, but they do not affect subsequent operations.

Next, we perform plane segmentation on the filtered point cloud. Using RANSAC, we randomly sample three points to fit a plane model and identify the one with the largest number of inliers (i.e., points within \SI{0.01}{\meter} of the plane). 
The inliers of this dominant (i.e., largest) plane are then extracted as the final planar point cloud.
\subsubsection{Hole Edge and Center Extraction\label{lidar_hole_center}}
After obtaining the planar point cloud of the calibration target, we apply voxel downsampling with a resolution of \SI{8}{\milli\meter} along all three axes. We then extract the edge points of the circular holes.
To facilitate this, the planar point cloud is rotated and aligned to $z = 0$ plane, resulting in a 2D-aligned point cloud. In a 2D plane, non-boundary points typically have neighbors evenly distributed in all directions, whereas edge points exhibit significant directional gaps in their local neighborhood. For each 2D point $\mathbf{p}_i$, the direction angles to its neighboring points $\mathbf{p}_j$ within a \SI{0.03}{\meter} radius are computed as follows:
\begin{align}
    \theta_j &= \text{atan2}\big( (\mathbf{p}_j - \mathbf{p}_i)_y,\ (\mathbf{p}_j - \mathbf{p}_i)_x \big)
\label{eq_theta}
\end{align}
These angles are sorted in ascending order, and the differences between adjacent angles are calculated in Eq.~\eqref{eq:delta_theta} to identify the largest angular gap. If the maximum gap exceeds \SI{25}{\degree}, the point is classified as an edge point.
\begin{equation}
\begin{aligned}
    \Delta \theta_k &= \theta_{k+1} - \theta_k, \,\, \text{for } k = 1, \dots, N{-}1 \\
    \Delta \theta_N &= \theta_1 + 2\pi - \theta_N
\end{aligned}
\label{eq:delta_theta}
\end{equation}

Edge points are then grouped using Euclidean clustering. To identify circular hole candidates and compensate for edge dilation caused by LiDAR spot spread, we perform ellipse fitting on each cluster of edge points using the general conic form:
\begin{equation}
Ax^2 + Bxy + Cy^2 + Dx + Ey + F = 0
\end{equation}
where $(A, B, C, D, E, F)$ are the parameters to be estimated. 
We adopt the direct least squares method proposed by \cite{fitzgibbon1996direct}, with an additional constraint to ensure that the fitted conic is an ellipse, i.e., $B^2 - 4AC < 0$.
The center of the fitted ellipse is then computed analytically as:
\begin{equation}
x_c = \frac{2CD - BE}{B^2 - 4AC}, \quad y_c = \frac{2AE - BD}{B^2 - 4AC}
\end{equation}
Given the geometric prior of circular holes, we validate each fitted ellipse using two criteria:  
1) Its semi-major axis length must differ from the known hole radius by less than \SI{4}{\centi\meter};  
2) Its eccentricity must be sufficiently low to indicate a near-circular shape. 
Ellipses meeting both criteria are accepted as valid circular hole extractions.

Finally, the hole center coordinates are transformed from the $z = 0$ plane to the original coordinate frame, yielding the point set $\mathcal{P}_L$.
\subsection{Registration\label{registration}}
Given two sets of circular hole centers extracted in the camera and LiDAR coordinate frames, the goal is to find a rigid transformation that minimizes the distances between corresponding points. Under the assumption of a one-to-one correspondence between points $\mathbf{p}_i^C \in \mathcal{P}_C$ and $\mathbf{p}_i^L \in \mathcal{P}_L$, the optimal transformation $\mathbf{T}_{CL}$ is estimated by minimizing the following least-squares error:
\begin{equation}
\frac{1}{4 \cdot N} \sum_{i=1}^{4 \cdot N} \left\| \mathbf{p}_i^C - \mathbf{T}_{CL} \mathbf{p}_i^L \right\|^2
\label{eq:register}
\end{equation}
This problem admits a closed-form solution via the Kabsch method, which estimates the rigid transformation using singular value decomposition (SVD).
The formulation naturally extends to the case of $N$ distinct captures (i.e., multi-scene joint calibration), each contributing 4 matching points and yielding a total of $4 \cdot N$ constraints. The problem remains a point-to-point registration task with known correspondences.
\section{Experiment}

We validate our calibration tool on three sensor configurations: OS1-128, Livox Avia, and Mid360, each paired with a wide-angle camera (MV-CS050-10UC CMOS sensor with an LM5JCM lens). Our sensor suite has been open-sourced on GitHub\footnote[3]{\url{https://github.com/xuankuzcr/LIV_handhold}}. For each configuration, we collect four pairs of accumulated LiDAR point clouds and corresponding camera images. The intrinsic parameters and distortion model of the camera are pre-calibrated.
\subsection{Consistency Evaluation\label{sec:consistent}}
\begin{figure}[htp]
	\begin{center}
		{\includegraphics[width=1\columnwidth]{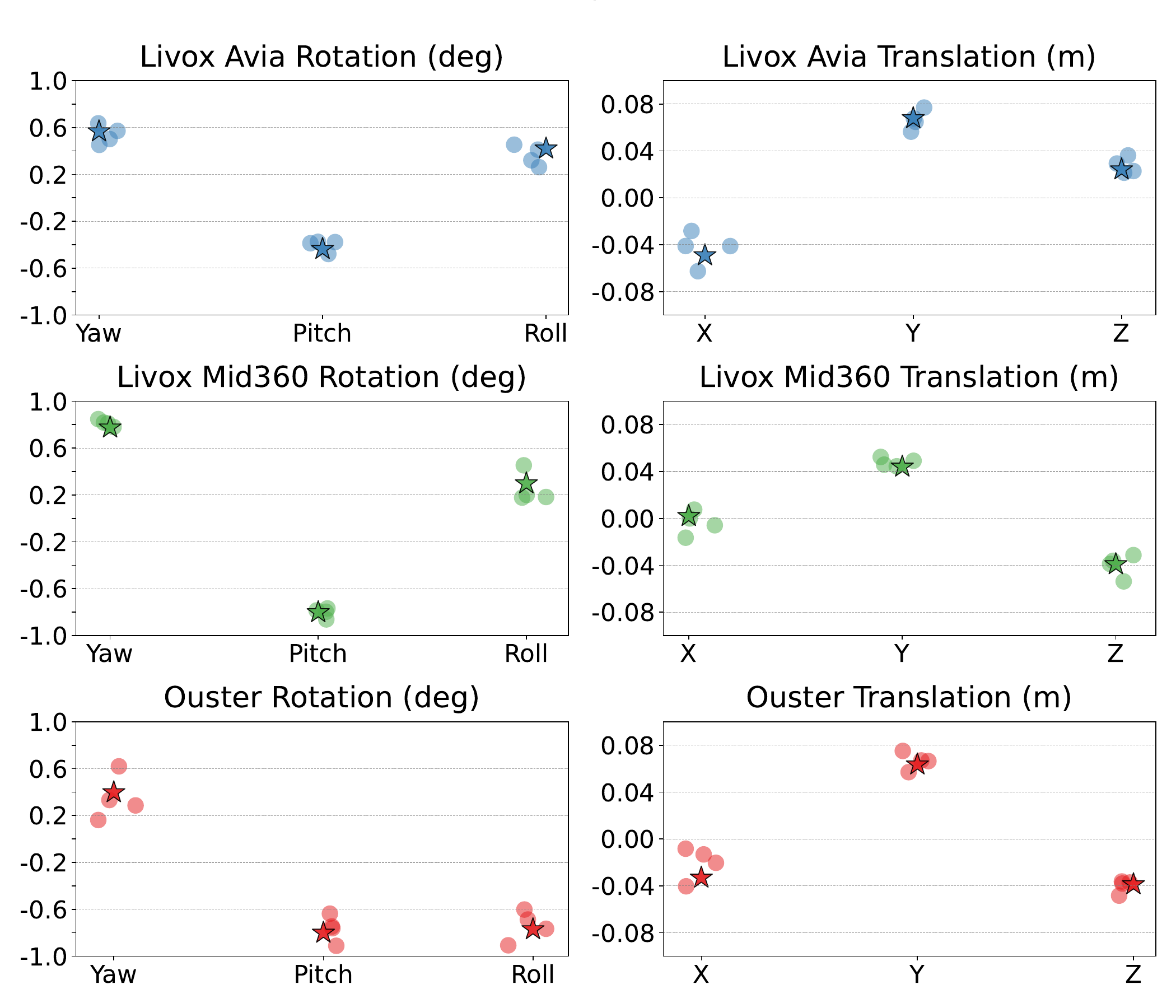}}
	\end{center}
      \vspace{-0.2cm}
	\caption{\label{fig:consist}The distribution of converged extrinsic parameters across six degrees of freedom for all data combinations. The nominal (integer) part of each value has been removed for clarity. Star markers denote the results from joint optimization over all four data pairs. To avoid overlap, a small random offset is added to the x-coordinates of the scatter points, while the star markers remain centered for ease of comparison.
}
\vspace{-0.1cm}
\end{figure}

To evaluate the consistency of the proposed calibration method, we perform five experiments for each sensor configuration: four runs using different combinations of three out of four data pairs for joint calibration, and one run using all four pairs for joint optimization. As shown in Fig. \ref{fig:consist}, the scatter plots illustrate the distribution of the converged extrinsic parameters for the Livox Avia, Mid360, and Ouster configurations. In each sensor configuration, the estimated extrinsic parameters between the camera and various LiDAR models exhibit low variance across all six degrees of freedom, demonstrating the consistency and robustness of FAST-Calib.


\begin{table}[htp]
    \centering
    \caption{Residual Comparison (in centimeters) Across Sensors and Algorithms}
    \label{tab:residual_comparison}
    \begin{tabularx}{\linewidth}{l|l|YYYYY}
        \toprule
        \textbf{Sensor} & \textbf{Algorithm} & \textbf{Run1} & \textbf{Run2} & \textbf{Run3} & \textbf{Run4} & \textbf{Run5} \\
        \midrule
        \multirow{2}{*}{Avia} 
            & Ours       & \textbf{0.15} & \textbf{0.22} & \textbf{0.38} & \textbf{0.65} & \textbf{0.25} \\
            & Velo2Cam   & 15.2 & 10.9 & 16.4 & 12.0 &  16.6 \\
        \midrule
        \multirow{2}{*}{Mid360} 
            & Ours       & \textbf{0.23} & \textbf{0.14} & \textbf{0.15} & \textbf{0.10} & \textbf{0.16} \\
            & Velo2Cam   & $\times$ & $\times$ & $\times$ & $\times$ & $\times$ \\
        \midrule
        \multirow{2}{*}{Ouster} 
            & Ours       & 0.29 & \textbf{0.17} &\textbf{0.21} & \textbf{0.30} & \textbf{0.21} \\
            & Velo2Cam   & \textbf{0.27} & 0.22 & 0.24 & 0.31 & 0.24 \\
        \bottomrule
    \end{tabularx}
    \begin{tablenotes}
        \item[1] $\times$ denotes the method totally failed.
    \end{tablenotes}
    \vspace{-0.2cm}
\end{table}
\subsection{Accuracy Evaluation}
\begin{figure}[t]
	\begin{center}
		{\includegraphics[width=1\columnwidth]{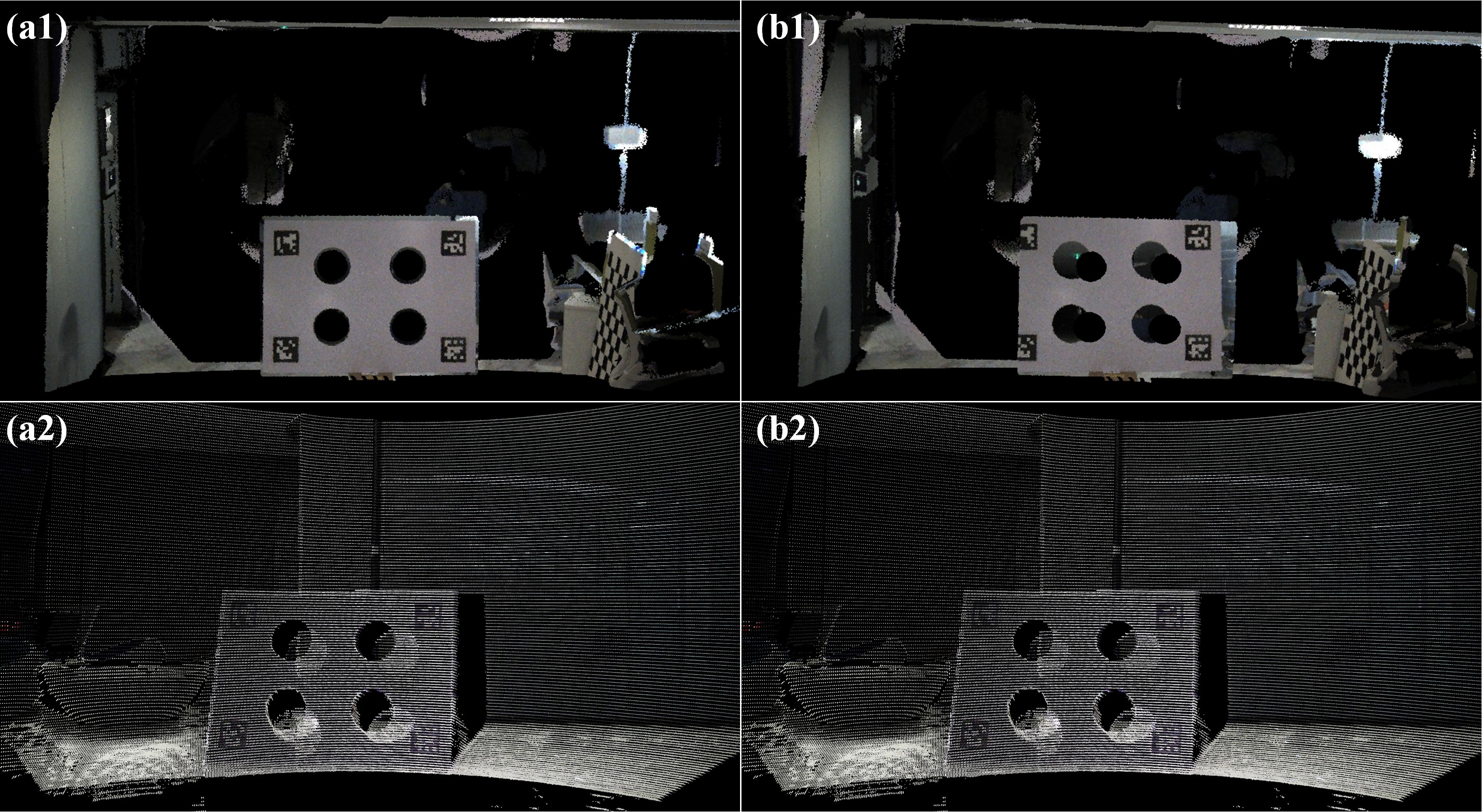}}
	\end{center}
      \vspace{-0.2cm}
	\caption{\label{fig:colored}(a) and (b) show the point clouds colored with the extrinsic parameters estimated by our method and Velo2Cam, respectively, with all data pairs used for joint calibration. (a1) and (b1) correspond to the Avia LiDAR, while (a2) and (b2) correspond to the Ouster LiDAR.
}
\vspace{-0.4cm}
\end{figure}

We conduct a quantitative comparison between our method and Velo2Cam\cite{beltran2022automatic} in terms of calibration accuracy. Following their setup, we employ the same 3D structured calibration target (see Fig. \ref{fig:target}). Velo2Cam estimates the centers of circular holes by detecting depth discontinuities along each ring, a technique that is theoretically applicable only to multi-line mechanical LiDARs. However, since the Livox Mid360 and Avia are equipped with 4 and 6 EEL laser emitters respectively, we adapt Velo2Cam to our data to obtain extrinsic calibration results. As described in Section \ref{sec:consistent}, we conduct five experiments for each sensor configuration. The residuals in \eqref{eq:register} are computed using the two sets of calibrated extrinsic parameters, and the quantitative results are presented in Table~\ref{tab:residual_comparison}.
``Run5'' denotes the result obtained using all available data pairs. Our method significantly outperforms Velo2Cam on solid-state LiDARs (i.e, Livox Avia and Mid360). For the Mid360 LiDAR, Velo2Cam fails completely in all five runs due to the extremely sparse and irregular distribution of its laser beams. For the Ouster LiDAR, our method achieves comparable calibration accuracy. In addition, Fig. \ref{fig:colored} shows the qualitative comparison of colored point clouds under the Livox Avia and Ouster sensor configurations.
In all cases, the point-to-point registration residuals remain below \SI{6.5}{\milli\meter}, demonstrating the high-precision alignment capability of FAST-Calib.
\subsection{Runtime Evaluation}
We evaluate the processing time for each calibration step. For each sensor configuration, the reported time corresponds to joint optimization using four data pairs. All experiments were conducted on a desktop PC equipped with an Intel i7-10700K CPU. Note that the reported time excludes camera intrinsic calibration and data collection. As shown in Table~\ref{tab:timing_analysis}, our method achieves a total processing time of less than \SI{0.7}{\second} across all sensor configurations.
The high efficiency of our method stems from several factors. By avoiding operations on dense point clouds and instead working with filtered point clouds downsampled to an \SI{8}{\milli\meter} resolution, computational overhead is significantly reduced. In addition, edge extraction is performed entirely in the 2D plane, enabling fast execution. Furthermore, multiple data pairs can be processed in parallel, leading to a total runtime that is approximately equal to that of processing a single pair.
\begin{table}[h]
    \centering
    \caption{Runtime Breakdown (seconds)}
    \label{tab:timing_analysis}
    \begin{tabular}{>{\raggedright\arraybackslash}p{4cm}ccc}
        \toprule
        \textbf{Calibration Stage} & \textbf{Avia} & \textbf{Mid360} & \textbf{Ouster} \\
        \midrule
        \textbf{LiDAR Data Processing}       & 0.633 & 0.675 & 0.647\\
        \quad ROI Extraction                 & 0.279 & 0.163 & 0.229\\
        \quad Hole Edge \& Center Extraction & 0.354 & 0.512 & 0.418\\
        \midrule
        \textbf{Camera Data Processing}      & 0.031 & 0.018 & 0.028\\
        \midrule
        \textbf{Registration}                & 2.3e-6 & 2.2e-6 & 3.1e-6 \\
        \midrule
        \textbf{Total}                       & 0.664 & 0.693 & 0.675\\
        \bottomrule
    \end{tabular}
\end{table}
\section{Conclusion}
We present FAST-Calib, a fast, accurate, and target-based extrinsic calibration tool for LiDAR–camera systems. The proposed method incorporates a novel algorithm for extracting the edges and centers of circular holes, which is independent of the LiDAR scanning pattern. It also supports joint optimization across multiple scenes. Both quantitative and qualitative experimental results demonstrate the consistency, robustness, and accuracy of our approach, as well as its applicability to mechanical and solid-state LiDARs. Moreover, the total processing time remains under one second across all sensor configurations.
With these capabilities, FAST-Calib effectively bridges the gap between production-line calibration and practical deployment in laboratory environments.



	
\section*{Acknowledgment}
    Special thanks to Jiaming Xu for his support, Haotian Li for the equipment.
	
	
	
	%
	
	\bibliographystyle{IEEEtran}
	\bibliography{paper}

\end{document}